\PassOptionsToPackage{breaklinks}{hyperref}
\documentclass[sigconf]{acmart}

\newcommand{\paperTitle}{Constructive Interpretability with CoLabel: Corroborative Integration, Complementary Features, and Collaborative Learning}
\newcommand{\paperKeywords}{Interpretability, Explainability, Vehicle Classification, VMMR}
\newcommand{\paperAuthors}{Abhijit Suprem, Sanjyot Vaidya, Suma Cherkadi, Purva Singh, Joao Eduardo Ferreira, Calton Pu}

\usepackage{endnotes,microtype,xspace,graphicx,fancyvrb,multirow}
\usepackage{pifont}
\usepackage{amsmath,amsopn}
\usepackage{listings}
\usepackage{subcaption}
\usepackage{mathrsfs}
\usepackage{booktabs}
\usepackage[labelfont=bf,font=small,textfont=md,belowskip=-5pt]{caption}
\usepackage[normalem]{ulem}
\useunder{\uline}{\ul}{}

\usepackage[breaklinks]{hyperref}
\hypersetup{%
    pdfauthor = {\paperAuthors},
    pdftitle = {\paperTitle},
    pdfkeywords = {\paperKeywords},
    bookmarksopen = {true},
    colorlinks=true,
    citecolor={urlcolor},
    linkcolor={linkcolor},
    urlcolor={citecolor},
    pdfborder={ 0 0 0 }
}

\usepackage[capitalize,noabbrev,nameinlink]{cleveref}

\newcommand{\squishitemize}{
\begin{list}{$\bullet$}
	{ \setlength{\itemsep}{0pt}
		\setlength{\parsep}{3pt}
		\setlength{\topsep}{3pt}
		\setlength{\partopsep}{0pt}
		\setlength{\leftmargin}{1.95em}
		\setlength{\labelwidth}{1.5em}
		\setlength{\labelsep}{0.5em} } }

\newcounter{Lcount}
\newcommand{\squishlist}{
	\begin{list}{\arabic{Lcount}. }
		{ \usecounter{Lcount}
			\setlength{\itemsep}{0pt}
			\setlength{\parsep}{3pt}
			\setlength{\topsep}{3pt}
			\setlength{\partopsep}{0pt}
			\setlength{\leftmargin}{2em}
			\setlength{\labelwidth}{1.5em}
			\setlength{\labelsep}{0.5em} } }
	
\newcommand{\squishend}{\end{list}}

\newcommand{\posthoc}{\textit{post-hoc}\xspace}

\usepackage{xstring}
\newcommand{\PP}[1]{
\vspace{2px}
\noindent{\bf \IfEndWith{#1}{.}{#1}{#1.}}
}

\def\equationautorefname~#1\null{Eq. ~(#1)\null}

\def\Snospace~{\S{}}

\newcommand{\sys}{\textsc{CoLabel}\xspace}
\newcommand{\sysnoatt}{\textsc{NoAtt}\xspace}
\newcommand{\sysmulti}{\textsc{CoLabel-MultiInput}\xspace}
\newcommand{\sysfusion}{\textsc{CoLabel-FusionOnly}\xspace}
\newcommand{\sysmatch}{\textsc{CoLabel-Match}\xspace}
\newcommand{\sysml}{\textsc{CoLabel-SMBL}\xspace}
\newcommand{\syscascade}{\textsc{CoLabel-2SC}\xspace}
\newcommand{\syscasmatch}{\textsc{CoLabel-2SC-Match}\xspace}
\newcommand{\corr}{\textsc{Corroborative Integration}\xspace}
\newcommand{\comp}{\textsc{Complementary Features}\xspace}
\newcommand{\coll}{\textsc{Collaborative Learning}\xspace}

\newcommand{\dcircle}[1]{\ding{\numexpr181 + #1}}

\newcommand{\zerodisplayskips}{%
  \setlength{\abovedisplayskip}{0pt}%
  \setlength{\belowdisplayskip}{0pt}%
  \setlength{\abovedisplayshortskip}{0pt}%
  \setlength{\belowdisplayshortskip}{0pt}}
\appto{\normalsize}{\zerodisplayskips}
\appto{\small}{\zerodisplayskips}
\appto{\footnotesize}{\zerodisplayskips}

\newcommand{\ie}{\textit{i}.\textit{e}.\xspace}
\newcommand{\eg}{\textit{e}.\textit{g}.\xspace}

\AtBeginDocument{%
  \providecommand\BibTeX{{%
    \normalfont B\kern-0.5em{\scshape i\kern-0.25em b}\kern-0.8em\TeX}}}

\begin{document}

\title{\paperTitle}

\author{Abhijit Suprem}
\email{asuprem@gatech.edu}
\affiliation{%
  \institution{Georgia Institute of Technology}
  \city{Atlanta}
  \state{GA}
  \country{USA}
}

\author{Sanjyot Vaidya}
\email{vaidya@gatech.edu}
\affiliation{%
  \institution{Georgia Institute of Technology}
  \city{Atlanta}
  \state{GA}
  \country{USA}
}

\author{Suma Cherkadi}
\email{suma@gatech.edu}
\affiliation{%
  \institution{Georgia Institute of Technology}
  \city{Atlanta}
  \state{GA}
  \country{USA}
}

\author{Purva Singh}
\email{purva@gatech.edu}
\affiliation{%
  \institution{Georgia Institute of Technology}
  \city{Atlanta}
  \state{GA}
  \country{USA}
}

\author{Joao Eduardo Ferreira}
\email{jef@ime.usb.edu.br}
\affiliation{%
  \institution{University of Sao Paolo}
  \city{Sao Paolo}
  \country{Brazil}}

\author{Calton Pu}
\email{calton@cc.gatech.edu}
\affiliation{%
  \institution{Georgia Institute of Technology}
  \city{Atlanta}
  \state{GA}
  \country{USA}
}

\renewcommand{\shortauthors}{Suprem, et al.}

\begin{abstract}    
  Machine learning models with explainable predictions are increasingly sought after,
  especially for real-world, critical applications
  that require bias detection and risk mitigation.
  Inherent interpretability, where a model is designed from the ground-up to
  provide prediction explanations, provide more faithful insights to model
  performance.
  In this paper, we present \sys, a constructively interpretable model
  that provides explanations rooted in the ground truth.
  We evaluate \sys for vehicle feature extraction, since these are crucial for re-id
  traffic monitoring and management, tracking, and vehicle make-model recognition (VMMR).
  Each task is a critical application that requires interpretable models.
  By construction, \sys provides VMMR along with a 
  composite of interpretable features.
  These interpretable features are based on interpretable annotations
  of the ground truth labels.
  \sys first performs corroborative integration to join multiple datasets
  that each have a subset of desired interpretable annotations.
  Then \sys uses decomposable branches to extract interpretable features
  and fuses them together for final predictions.
  During feature fusion, \sys performs harmonization of
  the complementary branches so that VMMR features can be projected
  to the same semantic space for classification.
  We show that \sys achieves superior accuracy to the state-of-the-art black-box models.
  Further, \sys provides explanations that help 
  in self-diagnosing and verifying predictions.  
  \end{abstract}

    

\maketitle

\section{Introduction}
\label{sec:intro}
Machine learning models that are interpretable and explainable are increasingly sought 
after in a wide variety of industry applications \cite{explainpersonreid,deepssh,explainability3,explainability2}. 
Explainable models augment the black-box of deep networks by providing insights into their 
feature extraction and prediction \cite{fergusvis}. 
They are particularly useful in real-world situations where accountability, 
transparency, and provenance of information for mission-critical human decisions
are crucial, such as security, monitoring, and medicine \cite{explainpersonreid,reidsurvey}. 
Interpretable indicates model features designed from the get-go to be human-readable. 
Explainable indicates \posthoc analysis of models to determine feature importance in prediction.

Inherently interpretable models \cite{interpretability0} are designed from the get-go to provide explainable results. 
This contrasts with \posthoc explainability, where a black box model is analyzed to obtain 
potential explanations for its decisions. 
Inherently interpretable models provide more faithful explanations \cite{interpretability0}, since these are directly 
dependent on model design. 
Such models also avoid pitfalls for \posthoc explainability such as unjustified counterfactual 
explanations \cite{interpretability2}. 
Inherently interpretable models have higher trust under adversarial conditions \cite{mythos} 
since their predictions can be directly tied to training ground truth through 
model-generated explanations.

\begin{figure}[t] 
  \centering \includegraphics[width=3.4in]{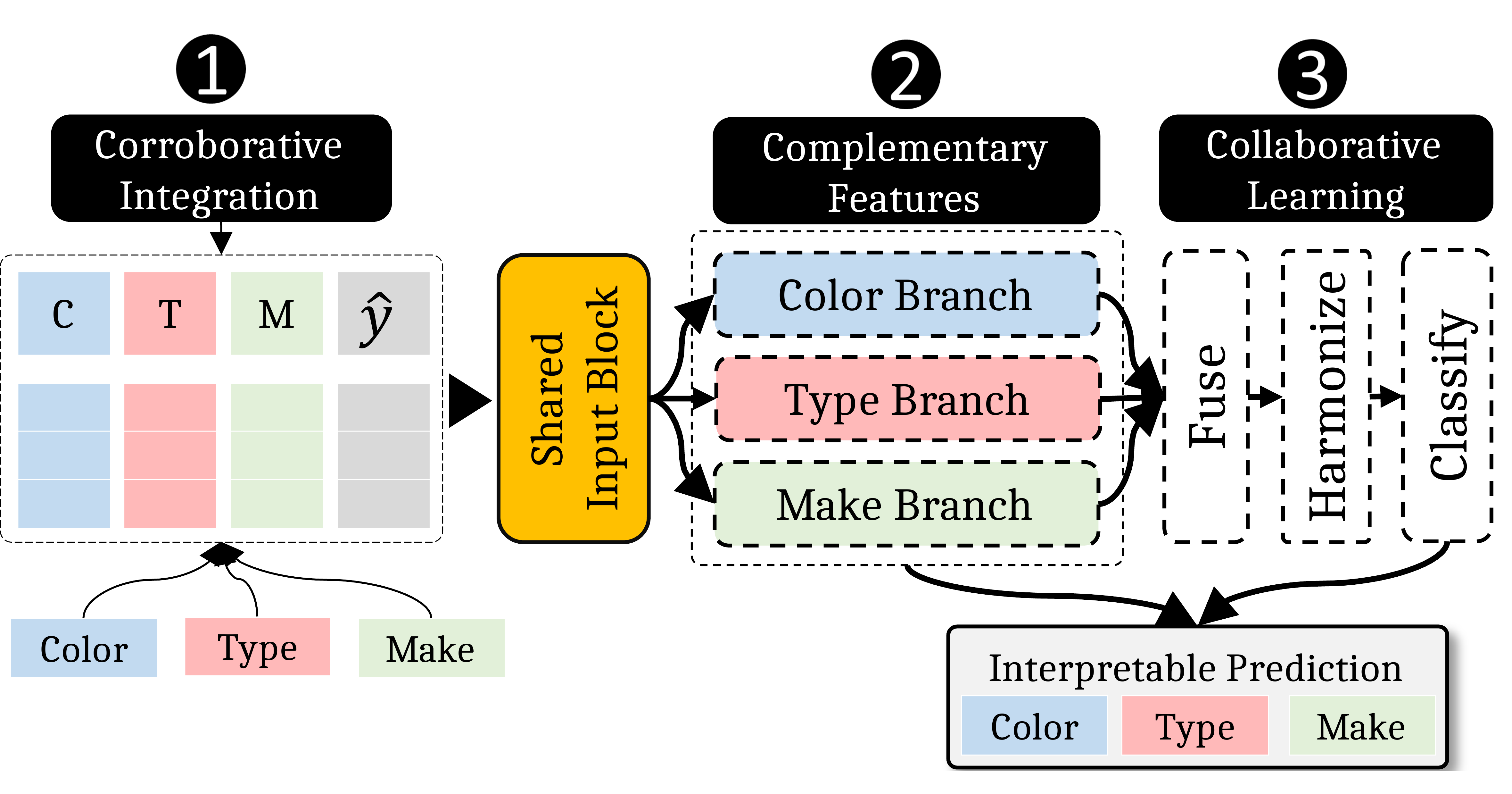}
\caption{\textbf{\sys Architecture and Dataflow}: \sys takes in multiple vehicle datasets, each with a subset of desired annotations.
\dcircle{1} \corr integrates annotations into a single training set.
\dcircle{2} \comp extracts interpretable features with complementary branches.
\dcircle{3} \coll fuses complementary features to yield interpretable predictions for vehicle classification.
 }
\label{fig:overview}
\end{figure}

There are several challenges, however, in building models that are inherently interpretable. 
There is no one-size-fits-all solution since interpretability is domain-specific \cite{interpretability0}. 
Existing datasets may not have completely interpretable annotations; instead, most datasets only 
have the ground truth annotations without interpretable subsets. 
For example, existing vehicle classification datasets label vehicle make and model, 
but not vehicle color, type, or decals \cite{datasetsurvey}. 
Furthermore, deep networks are biased during training towards strong signals, and 
may ignore more interpretable weaker signals. 
For example, person re-id models focus primarily on a person's shirt, and need guidance 
to focus on more interpretable accessories such as hats, handbags, or limbs \cite{explainpersonreid}.

\PP{CoLabel}
In this paper, we present \sys: a process for building inherently interpretable models. 
We use \sys to build end-to-end interpretable models that provide explanations rooted in 
the ground truth. 
By construction, \sys provides predictions along with a composite of interpretable features that comprise 
the prediction with a combination of \corr, \comp, and \coll.
We call this this approach to achieve interpretable models from design and implementation \textit{constructive interpretability}.

We demonstrate the inherent interpretability and superior accuracy of \sys in vehicle feature extraction,
an important challenge in monitoring, tracking, and surveillance applications \cite{mythos,interpretability0}.
%
Specifically, vehicle features are crucial for re-id, traffic monitoring 
and management, tracking, and make/model classification. 
Current state-of-the-art vehicle classification models employ black-box models.

These mission-critical applications require interpretable predictions for 
aiding human decisions, particularly for borderline cases where explanations 
aligned with human experience can benefit human decisions much more than 
algorithmic internal specifics. 
The goal of constructive interpretability is to design and build inherently 
interpretable models aligned with human understanding of applications.
This is where \sys comes in.
As an inherently interpretable model with
constructive interpretability in mind, \sys has state-of-the-art accuracy as well
as interpretable predictions.

We show \sys{'s} dataflow with respect to vehicle feature extraction in \autoref{fig:overview}. 
Our constructive interpretability approach for inherently interpretable models 
begins from selection of interpretation annotations 
for vehicle features: \textbf{color}, \textbf{type}, and \textbf{make}.
\sys uses these annotations to generate interpretable vehicle features. 
These features are usable in a variety of applications, such as vehicle make and model recognition (VMMR), re-id, tracking,
and detection \cite{datasetsurvey}.
In this work, we focus \sys on VMMR.
%

\PP{Dataflow}
Given our desired annotations of color, type, an make, as well as datasets that each carry a subset 
of these annotations, \sys{'s} dataflow involves the following three steps:

\dcircle{1}  \corr integrates multiple datasets and corroborates annotations
of 'natural' vehicle features across them. It then builds a robust training set 
with ground truth as well as interpretable annotations. 

\dcircle{2} The  \comp module extracts features corresponding to the interpretable annotations. 
The goal is to maintain interpretable knowledge when integrated.
Each complementary feature is extracted with its own branch in the \sys model. 
These features $x_{color}$, $x_{type}$, and $x_{make}$ are crucial for interpretable predictions.

\dcircle{3}  Finally, \coll harmonizes complementary features,
ensuring features from different branches can be fused more effectively.
%
With harmonization, branches collaborate to exploit correlations between 
complementary features.
During training, \sys backpropagates a harmonization loss $L_H$ on the error between prediction $y_i$ and ground truth $y$, 
as well as branch-specific losses $L_(B^(color,type) )$ on the errors between $x_(color,type)$ and ground truth annotations 
$y_(color,type)$. Simultaneously, $L_H^k$ improves feature fusion by ensuring branches collaborate on overlaps 
between complementary branches by exploiting correlations between complementary features.

\PP{Contributions}
We show that \sys can achieve excellent accuracy on feature extraction while simultaneously providing 
interpretable results by construction. 
\sys{'s} explanations align with human knowledge of vehicles, avoiding potential difficulties
of \posthoc explainability \cite{interpretability0,interpretability2,mythos}. 
The contributions are:

\squishitemize
\item \textbf{\sys}: Constructive interpretability approach to design and build an inherently interpretable vehicle feature extraction system by 
integrating diverse interpretable annotations that are aligned with human knowledge of vehicles.

\item \textbf{Model}: Experimental evaluation and demonstration of the superior accuracy and interpretability 
achieved by \sys 

\item \textbf{Loss}: A harmonization loss for fusion function to integrate complementary feature branches
to achieve high accuracy and faster convergence in \sys

\squishend

\section{Related Work}
\label{sec:related}
We first cover recent work in inherently interpretable models and \posthoc explanations. We will then cover vehicle feature extraction.

\begin{figure}[t] 
    \centering \includegraphics[width=3.4in]{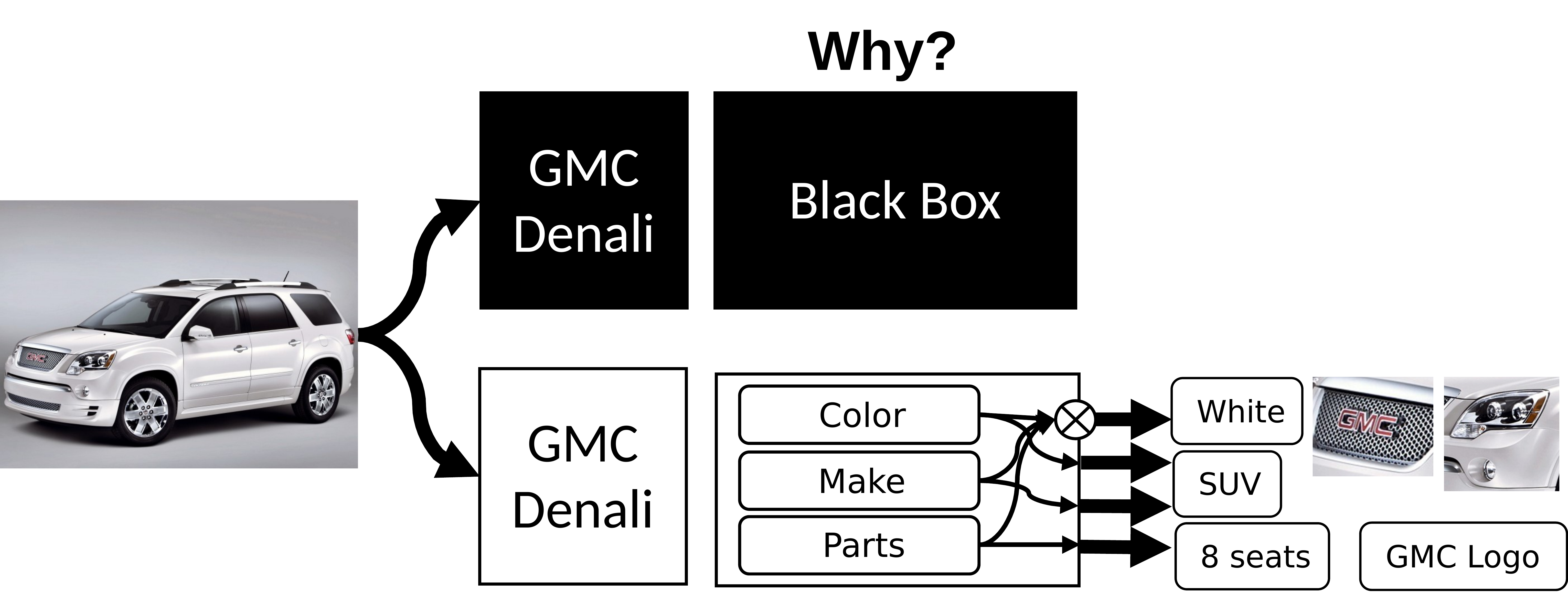}
  \caption{\textbf{Constructive Interpretability}: Models
  with constructive interpretability provide explanations for their predictions.
  This is achieved by incorporating
    human knowledge. Here, an interpretable model
    explains its prediction is based on vehcle color, type, seats, and vehicle part similarities.
   }
  \label{fig:inherent}
  \end{figure}

\subsection{Interpretability and Explainability}
\label{sec:interpexplain}

\PP{Interpretability}
Interpretable models are designed from the ground up to provide explanations
for their features.
Intuitively, interpretability is deeply intertwined with human 
understanding \cite{interpretability2}.
Models that are inherently interpretable directly integrate 
human understanding into feature generation.
Such models are more desired is mission-critical
scenarios such as healthcare, monitoring, and 
safety management\cite{interpretability0,interpretability1,mlbias1}. 
The prototype layers in \cite{explainability1,interpretability1} provide interpretable predictions: 
the layers compare test image samples to similar ground truth samples to provide
explanations of the model classification.
The decomposable approaches in \cite{cbc,credit} build component-classifiers that are 
integrated for the overall task, e.g. image and credit classification, respectively.
Like these, \sys is an inherently interpretable model design whose explanations 
are derived directly from training on annotated ground truth.

\begin{figure*}[t] 
  \centering \includegraphics[width=\textwidth]{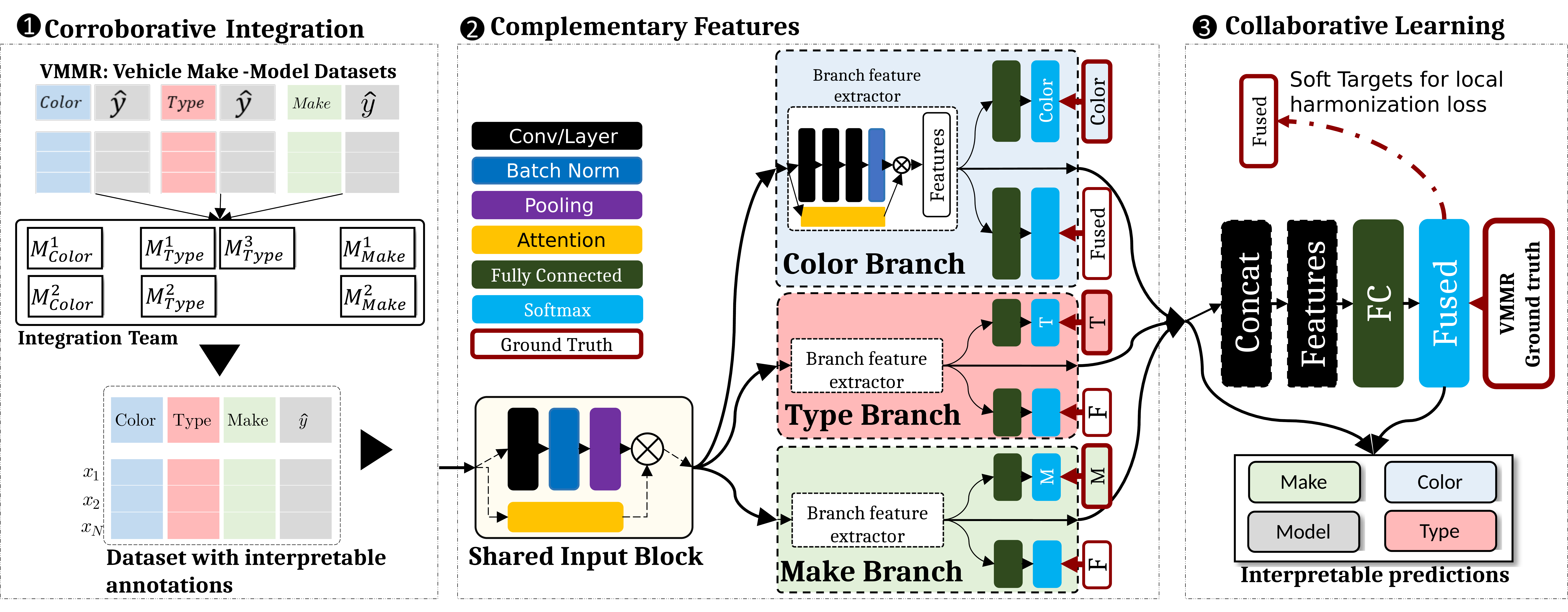}
  \caption{\textbf{\sys for VMMR}: 
  For VMMR, we use color, type, and make labels as our interpretable annotations.
  \dcircle{1} \textbf{\corr} combines various VMMR datasets, each with a subset of desired 
  interpretable annotations,  with a labeling team to generate a single dataset with all desired annotations (\autoref{sec:corr}).
  \dcircle{2} Then, \textbf{\comp} uses 3 branches to extract color, type, and make features. Each branch
  contains a feature extractor backbone. A dense layer converts features to branch-specific predictions. A
  second dense layer yields harmonization features (\autoref{sec:comp}).
  \dcircle{3} Finally, \textbf{\coll} fuses features for VMMR classification. Simultaneously, a harmonization loss
  ensures branch features collaborate on feature correlations (\autoref{sec:coll}).
  Predictions are combined from \coll and \comp to generate interpretable classification that correspond
  to annotations from \corr.
}
\label{fig:colabel}
\end{figure*}

\PP{Explainability}
Currently, most approaches perform \posthoc explanation in a bottom-up fashion, 
where an existing model's black-box is 'opened' \cite{sagemaker}. 
These include examining class activations \cite{gradcam0,pgan}, concept activation 
vectors \cite{explainability2}, neuron influence aggregation \cite{summit}, 
and deconvolutions \cite{fergusvis}. 
In \posthoc explanation, second model is used to model the original model's 
predictions by projecting model features along human-readable dimensions, if possible 
\cite{pgan,gradcam,summit,bluff}.
However, these approaches do not build interpretable models from the 
ground truth; they merely enhance existing models for explanation. 
For example, Grad-CAM's \cite{gradcam0} outputs are used with human labeling 
to determine 'where' and 'what' a model is looking at \cite{pgan,gradcam}.
Similarly, the embedding and neuron views in \cite{summit,bluff} make it 
easier to visually characterize an class similarity clusters. 
However, there are risks to explainability when it is disconnected from the \
ground truth \cite{interpretability0,interpretability2,mythos}. 
Such explanations may not be accurate, because if they were, the explainer model 
would be sufficient for prediction \cite{interpretability2}. 
Furthermore, interpretable models are usually as accurate as black-box models, with
the added benefit of interpretability, with several examples provided in \cite{rudin}.
Thus, the challenges of bottom-up approaches are bypassed with \textit{constructive interpretability},
since it is a top-down approach for interpretability, as in \autoref{fig:inherent}.
In constructive interpretability, inherently interpretable models 
are built with features aligned with human knowledge of application domains, 
forming intuitively understandable and interpretable models as in 
\cite{interpretability0,interpretability1,interpretability2,explainability1,cbc,credit}.

\subsection{Vehicle Feature Extraction}
\label{sec:features}

We demonstrate \sys{'s} interpretability with vehicle feature extraction. This encompasses
several application areas, from VMMR\cite{cooccur,vmmrcmp,vmmr}, re-id \cite{reid1,reid2,reid3,glamor}, 
tracking \cite{track1,track2,track3,track4}, and vehicle detection\cite{boxcars,detect}. 
We cover recent research on feature extraction for these application areas.

Hu et al. presented a framework for identifying vehicle parts for extracting more 
discriminative representations without supervision. 
CNN and SSD models are used together for logo 
detection in high resolution images\cite{logo}. 
Logo-Net \cite{logonet} uses such a composition to improve logo detection and classification. 
Wang et. al. \cite{oife} develop an orientation-invariant approach that uses 20 engineered 
keypoint regions on vehicles to extract representative features. 
Liu et. al. propose RAM \cite{ram}, which  has sub-models that each focus on a different 
region of the vehicle's image, because different regions of vehicles have 
different relevant features. 

Additionally, there have been recent datasets with varied annotations for feature extraction.
Boukerche and Ma \cite{datasetsurvey} provide a survey of such datasets for feature extraction.
Yang et. al. \cite{compcars} propose a part attributes driven vehicle model recognition 
and develop the CompCars dataset with VMMR labels. 
BoxCars116K \cite{boxcars} provides a dataset of vehicles annotations with type,
and uses conventional vision modules for vehicle bounding box detection.

\PP{Summary}
Since each application area for vehicle features remains sensitive and mission-critical, interpretable features
are crucial for deployment.
The above approaches have improved on feature extraction. We augment them with \sys to demonstrate 
interpretability. 
We will further show that such inherently interpretable models offer additional avenues for increasing
model accuracy.

\section{CoLabel}
\label{sec:colabel}

\sys, our approach for interpretable feature extractions (\autoref{fig:colabel}). 
We have given an overview of \sys{'s} dataflow in \autoref{fig:overview}. 
Here, we describe \sys{'s} components in details.
We first describe \dcircle{1} \corr in \autoref{sec:corr}.
Then we present \dcircle{2} \comp for interpretable annotations in \autoref{sec:comp}.
Finally, we present \dcircle{3} \coll in \autoref{sec:coll}, where \sys fuses 
complementary features for interpretable predictions. 
We implement \sys for vehicle feature extraction, which has a need for interpretability 
in a variety of mission-critical applications in traffic management, safety monitoring, 
and multi-camera tracking. 
Specifically, we apply \sys for interpretable vehicle make and model recognition (VMMR), 
where the task is to generate features from vehicle images to identify the make and model.

\subsection{Corroborative Integration}
\label{sec:corr}

\begin{figure}[t] 
    \centering \includegraphics[width=3.4in]{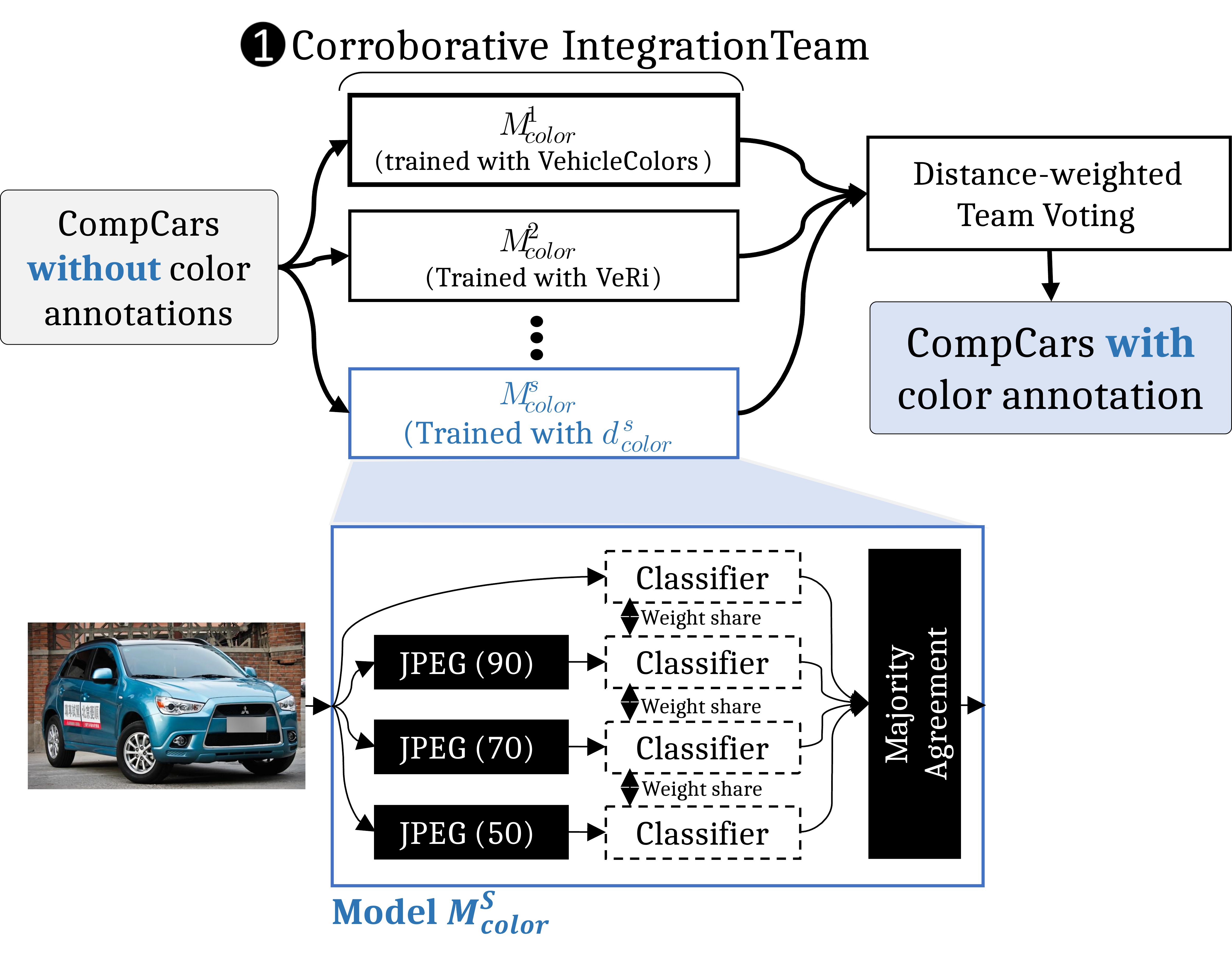}
  \caption{\textbf{\corr}: Given a desired annotation $k$ (e.g. color), a subset $d^s_{color}\in D$ of 
  VMMR datasets contains this annotation. \corr employs a team of labeling models $M^s_{color}$, each 
  trained on a corresponding $d^s_{color}$. Each model employs JPEG compression ensemble to improve labeling
  agreement. Using \corr, \sys can label the remaining $D-s$ datasets with the color annotations.
   }
  \label{fig:corr}
  \end{figure}

%
Constructive interpretability starts from a judicious decomposition 
of the application domain into interpretable annotations. 
For vehicle classification with \sys, we identified three interpretable annotations 
for model training in addition to VMMR labels: vehicle \textbf{color}, \textbf{type}, 
and \textbf{make}.
Vehicle color is the overall color scheme of a vehicle. 
Type is the body type of the vehicle, such as SUV, sedan, or pickup truck. 
Finally, make is the vehicle brand, such as Toyota, Mazda, or Jeep.
We select these annotations as they are broadly common across vehicle feature 
extraction datasets \cite{datasetsurvey}.

One advantage of \sys is a very inclusive classifier training process. 
Of the several VMMR datasets (discussed in \autoref{sec:datasets}),
each has a subset of the three desired interpretable annotations. 
CompCars \cite{compcars} labels make, model and type. 
VehicleColors \cite{vcolors} labels only the colors. 
BoxCars116K \cite{boxcars} labels make and model.
\sys is capable of integrating the partial knowledge in all of labeled data sets
with \corr.

\PP{Labeling Overview}
So, \sys uses \corr to `complete' partial annotations: in this case by adding 
color labels to each image in CompCars. 
Let there be $k$ interpretable annotations, and a set $D$ of datasets, all of which contain some subset of 
$k$ annotations/labels $\hat{y}_k$. 
Datasets in $D$ also contain the overall ground truth label $\hat{y}$, \ie VMMR. 
For \sys, $k=3$: color ($\hat{y}_{color}$), type ($\hat{y}_{type}$) and make ($\hat{y}_{make}$).
We show \corr  for a single interpretable annotation ($\hat{y}_{color}$) in \autoref{fig:corr}. 

So, a subset $\{d_{color}\}_s\in D$ contains desired annotation $\hat{y}_{color}$. 
We train a set of models ${M_{color}}_s$, one for each of the $s$ datasets in $\{d_{color}\}_s$, 
with the datasets' corresponding $\{\hat{y}_{color}\}_s$ as the ground truth. 
Then, we build an team of $\{M_{color}\}_s$ to label the remaining datasets in $D$ 
without color annotation, \eg the $\hat{y}_{color}$-unlabeled subset $\{d_{color}^*\}_{|D|-s}$, 
with weighted voting.
Since $d_{color}^*$ has a subset of desired annotations, we call it a partially
unlabeled dataset; this subset is the complement of labeled subset $d_{color}$.

\PP{Labeling Team}
Team member votes are dynamically weighted for each partially unlabeled dataset $d_{color}^*$. 
First, we partition the unlabeled dataset into $l$ 
clusters with KMeans clustering.
We also partition each team member's training dataset into $l$ clusters.
Then, for each cluster in the unlabeled dataset, we find the 
nearest training dataset using cluster
overlap as the guiding metric.
The corresponding team member is used for labeling.
We compute overlap between unlabeled dataset clusters and team
member training data using the O-metric \cite{ometric} for point-proximity calculation.

O-metric calculates overlap between 2 n-dimensional clusters using 
some distance metric; in our case, we use cosine distance.
O-metric works as follows: given an unlabeled cluster $U_i$ and a 
labeled cluster from any team member training data $T_j$,
we compute the fraction of points in $U_i$ that are closer to 
points in $T_j$.
So, for each $u\in U_i$, we calcualte the point proximity value with:

$$
O_U(u) = conspecific(u) / heterospecific (u)
$$

Here, $conspecific()$  calculates the distance between $u$ 
to the nearest point in $U_i$.
Conversely, $heterospecific()$ computes the distance between $u$
to the nearest point in $T_j$.
Then, we compute the proximity ratio to determine overlap of $T_j$ in $U_i$:

$$
p_U = \frac{|O_U > 1|}{|U_i|}
$$

The value of $p_U$ is bounded in $[0,1]$. As $p_U$ gets closer to 1,
this indicates most points in $U_i$ are closer to some point in $T_j$
than in its own cluster $U_i$.
We can compute the pairwise cluster overlap between every unlabeled cluster
to every labeled cluster.
For each unlabeled cluster, we only need to take the labeled cluster
with the highest overlap.
This is similar to the dataset distance approaches in \cite{odin, kmp, ometric}.
%


Each trained model generalizes to the task for cross-domain datasets, as we will show in \autoref{sec:correval}.
However, models still encounter edge cases due to dataset overfitting \cite{cross}. 
We address cross-domain performance deterioration with early stopping, a JPEG compression ensemble,
and greater-than-majority agreement among models.

\PP{Early Stopping}
During training of each model in $\{M_{color}\}_s$, we compute validation accuracy over the validation sets 
in $\{d_{color}\}_s$. 
With early stopping tuned to cross-dataset validation accuracy, we can ensure models do not overfit to their own training dataset.

\PP{JPEG Compression}
During labeling of any $d_{color}^*$, each labeling model $M_{color}$ takes 4 copies of an unlabeled image.
The first is the original image. 
Each of the three remaining copies is the original image compressed using the JPEG protocol, with 
quality factors of 90, 70, and 50.
We use the majority predicted label amongst the four copies as the model's final prediction.
This is similar to the 'vaccination' step from \cite{shield}, where JPEG compression removes
high-frequency artifacts that can impact cross-dataset performance.

Using JPEG compression essentially defends against adversarial attack.
In this case, the `adversarial attack' is potential label changes due to poor 
local coverage of the model's embedding space.
Neural networks are typically smooth around their embedding space \cite{nice, liger}.
The smoothness is computed with the ratio of the change in embedding with respect
to change in input.
If this ratio is larger than 1, this means nearby points in the input space
are not clustered together in the embedding space.
Conversely, a ratio smaller than 1 indicates nearby points are clustered
together: a desirable property since nearby points are typically 
in the same class \cite{liger, nice, midas}.
However, calculation of this ratio is a computationally expensive process,
since it requires either clustering the input and embedding space \cite{liger}
or generating perturbations for each input and comparing them to embeddings \cite{midas}.
With JPEG compression, we use a well-known technich to slightly perturb
the input image, and check if the labels change.
If labels change, this indicates the local region around the input image
is not smooth, since the model is changing labels due to minor perturbations.
In this case, we can use the majority label as a proxy for directly computing 
the smoothness.

\PP{Team Agreement}
If an image has no majority label from a model's JPEG ensemble, we discard that model's label.
Further, if we only have $<50\%$ of models in the labeling team after discarding 
models without JPEG ensemble majority label, we leave the annotation for that image blank.
This is similar to a reject option that directly uses an estimate for 
smoothness. 
In our case, computing the JPEG compressions and evaluating 
on 4 images is significantly faster than the clusterign plus smoothness
estimation method in \cite{liger, midas}.

\PP{Summary}
Using these steps, we can label most partially unlabeled samples datasets 
without the desired annotation.
We evaluate labeling accuracy given these strategies in \autoref{sec:correval}, where we test on 
held-out labeled datasets. 
Our results show average accuracy improvement of almost 20\% from 0.83 to 0.98 for held-out 
unlabeled samples when we use labeling teams with early stopping for team members and 
JPEG compression ensemble for each model. 

\subsection{Complementary Features}
\label{sec:comp}

\sys{'s} next step is \comp extraction, which propagates the three interpretable annotations
from \corr for feature extraction.
Here, \sys explicitly learns to extract interpretable features from the interpretable 
annotations generated by \corr. 
\comp comprises of 2 stages: a shared input stage, followed by $k=3$ complementary 
feature branches. 
The shared input stage performs shallow convolutional feature extraction.
These features are common to each branch's input. 
Then, the complementary branches extract their annotation-specific interpretable features 
and propagate them to \coll.

\PP{Shared Input Block}
Multi-branch models often use different inputs for each branch \cite{mtml,vami}. 
\sys uses a shared input block to create common input to each branch to improve feature fusing in \coll. 
Feature fusing requires integration of branches that carry different semantics and scales.
This is accomplished with additional dense layers after concatenation \cite{oife}, longer training to ensure convergence
or appropriate selection of loss functions \cite{mtml}.
\sys{'s} branches perform complementary feature extraction, where the features 
are semantically different.
So, we use a shared input block to ensure branches have a common starting point in shallow
convolutional features and use attribute features from those layers \cite{explainpersonreid,fclayers}. 
It consists of early layers in a feature extractor backbone such as ResNet, 
along with attention modules (we use CBAM \cite{cbam}). 
For \sys, we use the first ResNet bottleneck block in the shared input layer, and use the remaining
bottleneck blocks in the branches.
We show the the impact of the shared input block for training in \autoref{sec:compeval}.

\PP{Complementary Feature Branches}
Each interpretable annotation from \dcircle{1} \corr is matched to a corresponding branch in \sys. 
We describe a single \comp branch here.
The features $x_{shared}$ from the shared input block are passed through a feature extractor backbone 
comprising of conv layers, normalization, pooling, and CBAM.
This yields intermediate features $x_k$, \eg $x_{color}$ for the color branch. 
A fully connected layer $F_{color}$ projects $x_{color}$ to predictions $y_{color}$. 
A second fully connected layer $F_{color-fused}$ projects $x_{color}$ to tentative fused predictions $y_{color-fused}$
These tentative fused predictions are only used during training to improve feature harmonization in \coll.
We defer their discussion to \autoref{sec:coll}. 
Finally, $x_{color}$ is also sent to \coll for fusion with the other branch features  $x_{type}$ and $x_{make}$.

\PP{Training}
During training, we compute 2 local losses to train the branch feature extractors.
$L_B^k$ is the branch specific loss for branch $k$, computed as the cross-entropy loss between $y_k$ and $\hat{y}_k$:

\begin{equation}
    L_B^{color}=\mathcal{H}(y_{color}, \hat{y}_{color})
    \label{eq:branch}
\end{equation}

\noindent
$L_H^k$, the local harmonization loss, is discussed in \autoref{sec:coll}.

\PP{Impact of Complementary Features}
\corr generates interpretable features in $x_{color}$.
As such, for any prediction, we can decompose \sys into the $k$ complementary branches
and explain predictions with the component annotations.
For prediction errors, \sys provides provenance of its classification, so that the specific branch
that caused the error can be updated.
Further, the interpretable branches are extensible: any new desired annotations need
only be added to the training set with \corr. 
Subsequently, \comp will deploy a branch to generate interpretable features for the corresponding
annotations.
We discuss the impact of \comp in accuracy and interpretability in \autoref{sec:compeval}
\subsection{Collaborative Learning}
\label{sec:coll}

\begin{figure}[t] 
    \centering \includegraphics[width=3.4in]{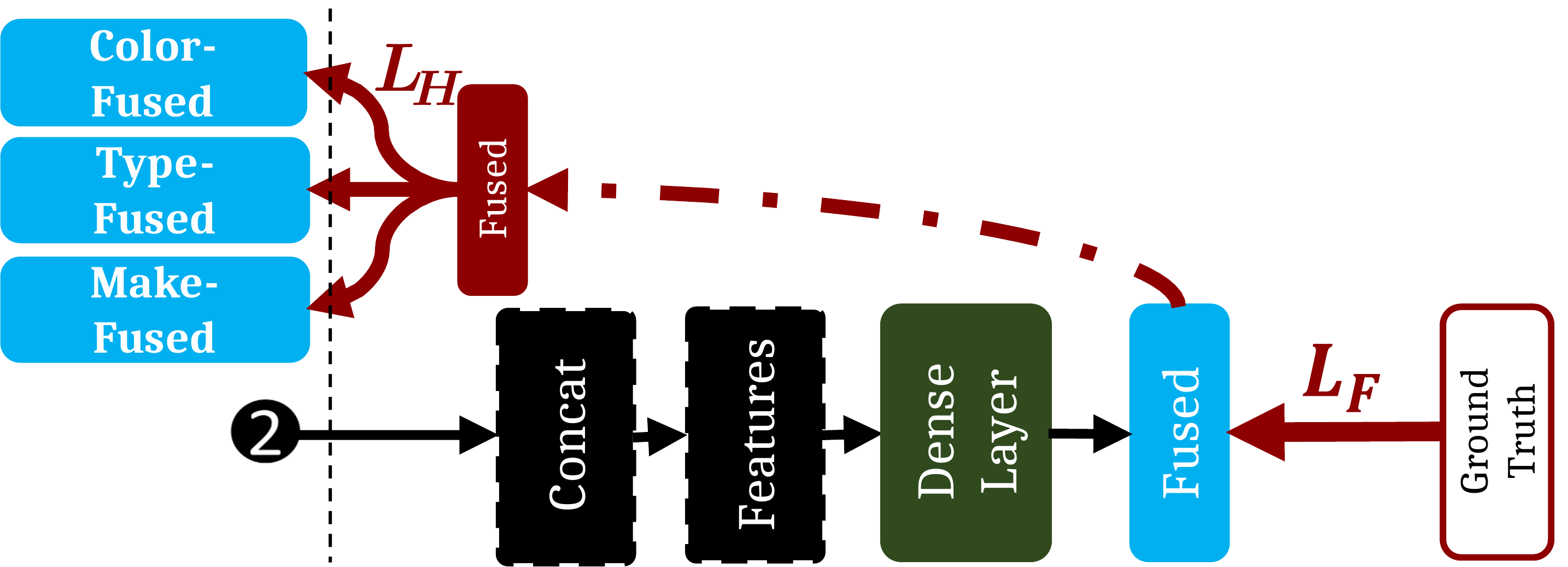}
  \caption{\textbf{\coll}: Features from \comp are fused with concatenation. 
  Branches collaborate on feature correlations and interdependencies with 
  the harmonization losses $L_H^k$. 
  $L_H^k$ uses the fused features as soft-targets for $x_{k-fused}$ 
  from \autoref{sec:comp}. 
  A fused loss $L_F$ on the cross-entropy loss between VMMR targets
  and fused prediction also backpropagates over the entire model.
   }
  \label{fig:coll}
\end{figure}

Finally, \sys performs feature fusion to generate interpretable predictions. 
The branches of \comp yield their corresponding features $x_{color}$, $x_{type}$, and $x_{make}$. 
These are concatenated to yield fused features $x_F$. 
A fully connected layer projects $x_F$ to final predictions $y_F$.
\sys is trained end-to-end with the cross-entropy loss between predictions
and ground truth:

\begin{equation}
    L_F=\mathcal{H}(y_F,\hat{y})=-\frac{1}{N_B}\sum^{N_B} p_i^{y_F} \log (p_i^{\hat{y}})
    \label{eq:lf}
\end{equation}

\PP{Local Harmonization Loss}
\sys employs a local harmonization loss for each branch in \comp. 
Since the branches are extracting complementary features, we need a way for branches to 
exploit correlation and interdependency between features. 
We accomplish this with weak supervision on the branch features using the fused feature predictions $y_F$.
Intuitively, we want branches to agree on the overall VMMR task.
So, branch features should also accomplish VMMR, in addition to their branch-specific annotation.
Using this fused-feature knowledge distillation, we ensure that branch features harmonize 
on the final VMMR prediction labels $y_F$.
In effect, $y_F$ is a soft target for each branch.  
The local harmonization loss is computed for each branch as the cross-entropy loss
between the tentative fused predictions $y_{k-fused}$ from \comp and the 
final fused predictions $y_F$. For the color branch:

\begin{equation}
    L_H^{color}=\mathcal{H}(y_{color-fused},y_F)
    \label{eq:lharm}
\end{equation}

\PP{Losses}
\sys employs 3 losses during training, shown as red arrows in \autoref{fig:colabel} and \autoref{fig:coll}. 
The fused loss $L_F$ in \autoref{eq:lf} backpropagates through the entire model. 
\sys{'s} branches are trained with branch annotation loss $L_B^k$ and local harmonization loss $L_H^k$:

\begin{equation}
    \begin{aligned}
        L_{k} & =   L_B^k + L_H^k = \mathcal{H}(y_{k}, \hat{y}_{k}) + \mathcal{H}(y_{k-fused},y_F) \\
              & = -\frac{1}{N_C}\sum^{N_C} p_i^{y_k} \log p_i^{\hat{y}_k} - \frac{1}{N_B}\sum^{N_B} p_i^{y_{k_fused}} \log p_i^{y_F}
    \end{aligned}
    \label{eq:lbranch}
\end{equation}

Here, $C$ is the subset of mini-batch $B$ that has the annotations for $x_k$. 
We need this because during \corr, \sys leaves unlabeled samples without team agreement as unlabeled.
These unlabeled samples can be processed under an active learning scheme.
For this paper, we compute loss using the subset of samples for which the annotation is known.
\section{Results}
\label{sec:eval}

Now, we show the effectiveness and interpretability of \sys. 
First, we will cover the experimental setup. 
Then we will evaluate each of \sys{'s} components and demonstrate efficacy of design choices.
Finally, we demonstrate interpretability as well as high accuracy with the end-to-end model for VMMR.

\subsection{Experimental Setup}
\label{sec:datasets}
We cover system setup, as well as datasets.

\PP{System Details}
We implemented \sys in PyTorch 1.4 on Python 3.8. 
For our backbones, we use ResNet with IBN \cite{baseline}, with pretrained ImageNet weights. 
Experiment are performed on a server with NVIDIA Tesla P100, and an Intel Xeon 2GHz processor.

\begin{table}[t]
    \caption{\textbf{Datasets}: We use the \textbf{boldfaced} datasets in our final evaluations. 
    They are partially annotated. 
    To complete the \textit{No} annotations, 
    we use the \underline{underlined} datasets for each annotation in
    \corr.}
    \label{tab:datasets}
    \begin{tabular}{lrrrr}
    \hline
    Dataset              & Make         & Model        & Color       & Type        \\ \hline
    \textbf{CompCars}\cite{compcars}    & \textbf{Yes} & \textbf{Yes} & \textit{No} & {\ul Yes}   \\
    \textbf{BoxCars116K}\cite{boxcars} & \textbf{Yes} & \textbf{Yes} & \textit{No} & \textit{No} \\
    \textbf{Cars196}\cite{cars196}     & \textbf{Yes} & \textbf{Yes} & \textit{No} & \textit{No} \\
    VehicleColors\cite{vcolors}        & No           & No           & {\ul Yes}   & No          \\
    VeRi\cite{veri}                 & No           & No           & {\ul Yes}   & {\ul Yes}   \\
    CrawledVehicles (ours)      & No           & No           & {\ul Yes}   & {\ul Yes}   \\ \hline
    \end{tabular}
    \end{table}

\PP{Datasets}
We use the following datasets:
\textbf{CompCars}\cite{compcars}, \textbf{BoxCars116K}\cite{boxcars}, 
\textbf{Cars196}\cite{cars196}, \textbf{VehicleColors}\cite{vcolors}, and 
\textbf{VeRi}\cite{veri}. 
We also obtained our own dataset of vehicles labeled with color and type 
annotations using a web crawler on a variety of car-sale sites, called
\textbf{CrawledVehicles}. 
Datasets are described in \autoref{tab:datasets}.

We use CompCars, BoxCars116K, and Cars196 for end-to-end evaluations. 
Their annotations are incomplete, since none contain all three desired annotations.
We complete the ground truth for these datasets with \corr.

\subsection{Corroborative Intergration}
\label{sec:correval}

\begin{table}[t]
\small
    \caption{\textbf{Color-CM}: Accuracy of Color-CM teams in labeling held-out datasets with color annotations. For each column's evaluation, the team member models are trained with the datasets of the other 2 columns.}
    \label{tab:colorcm}
    \begin{tabular}{@{}rrrr@{}}
    \toprule
    \multicolumn{1}{l}{}        & \multicolumn{1}{r}{VehicleColors} & \multicolumn{1}{r}{VeRi} & \multicolumn{1}{r}{CrawledVehicles} \\ \midrule
    \multicolumn{1}{l}{Initial} & 0.87                              & 0.84                         & 0.86                                \\ \midrule
    +Early Stop                 & 0.89                              & 0.9                          & 0.92                                \\
    +Compression (90)           & 0.91                              & 0.91                         & 0.92                                \\
    +Compression(90, 70, 50)    & 0.94                              & 0.93                         & 0.94                                \\ \midrule
    +Labeling Team              & 0.95                              & 0.95                         & 0.96                                \\
    +Agreement                  & 0.98                              & 0.97                         & 0.98                                \\ \bottomrule
    \end{tabular}
\end{table}

\begin{table}[t]

    \caption{\textbf{Type-CM}: Accuracy of Type-CM team in held-out dataset with type annotations.}
    \label{tab:typecm}
    \begin{tabular}{@{}rrrr@{}}
    \toprule
    \multicolumn{1}{l}{}        & CompCars & CrawledVehicles & VeRi \\ \midrule
    \multicolumn{1}{l}{Initial} & 0.86     & 0.88            & 0.86 \\ \midrule
    +Early Stop                 & 0.89     & 0.91            & 0.89 \\
    +Compression (90)           & 0.91     & 0.91            & 0.91 \\
    +Compression(90, 70, 50)    & 0.93     & 0.94            & 0.94 \\ \midrule
    +Labeling Team              & 0.96     & 0.95            & 0.95 \\
    +Agreement                  & 0.98     & 0.97            & 0.98 \\ \bottomrule
    \end{tabular}
\end{table}

CompCars, Cars196, and BoxCars116K are missing annotations from our desired 
interpretable annotation list of make, color, and type (see \autoref{tab:datasets}). 
We use \corr to augment these datasets.
Specifically, we use VehicleColors, CrawledVehicles, and VeRi to label
Cars196, BoxCars116K, and CompCars with color annotations.
Then, we use CompCars and CrawledVehicles to label Cars196 and 
BoxCars116K with type annotations.

\begin{figure}[t] 
    \centering \includegraphics[width=3.4in]{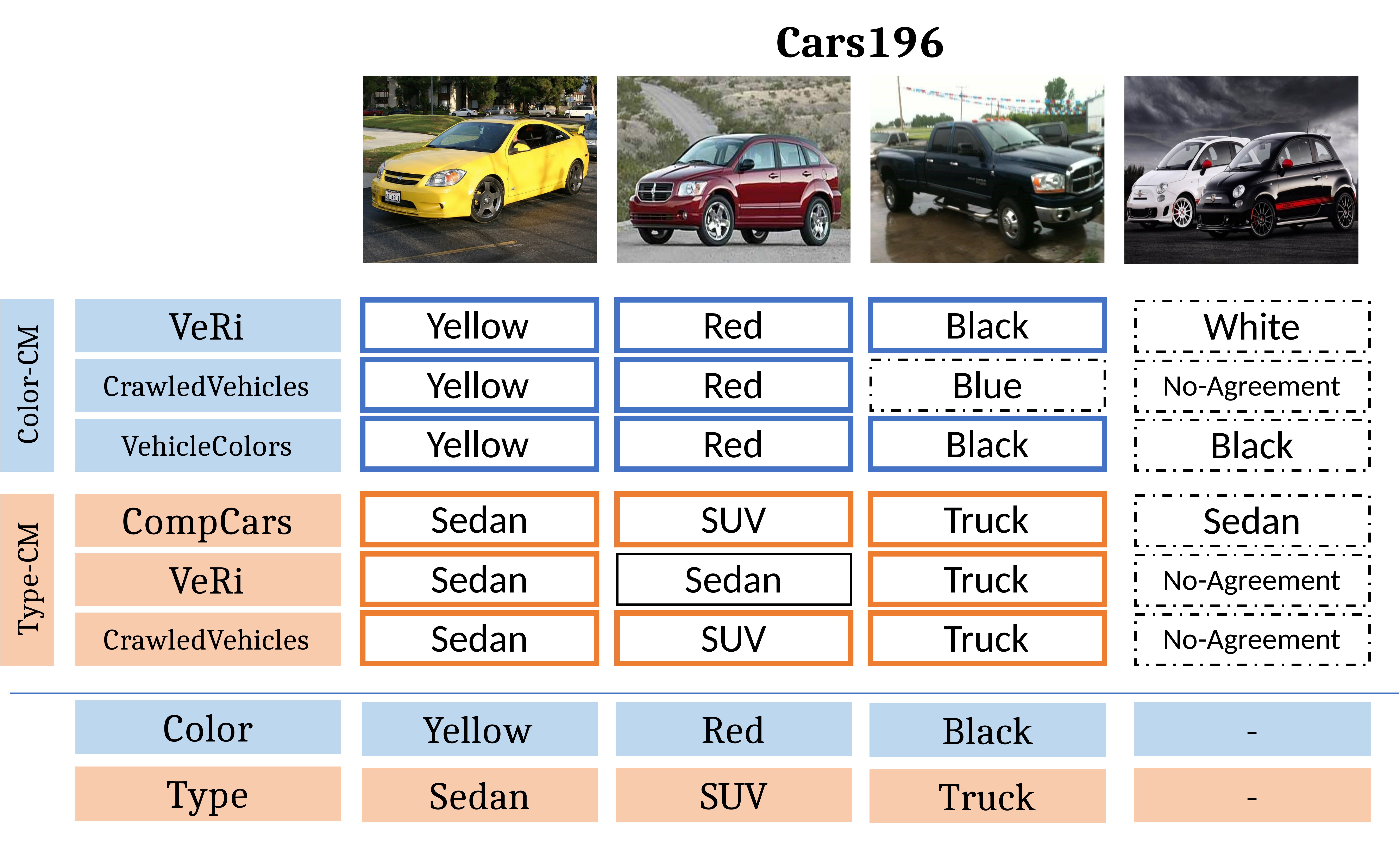}
  \caption{\textbf{Labeled Cars196 Images}: Using \corr, we can assign
  color and type annotations to Cars196 images.
  For some images with multiple vehicles or occlusions, we leave blank annotations
  if \corr cannot find agreement on the labels.
   }
  \label{fig:corrimages}
  \end{figure}

\PP{Color Model (Color-CM)}
Color-CM is a team of 3 models, where each model is trained with VehicleColors, CrawledVehicles, 
or VeRi, respectively.
During training of each model, we use horizontal flipping, random erasing, 
and cropping augmentations to improve training accuracy.
For the JPEG compression ratios of each model, we test with 2 schemes:
(a) with a single additional ratio of quality factor 90, and (b) 3
additional ratios with quality factors 90, 70, and 50. 
We use majority voting from the JPEG compression ensemble. 
Then with majority weighted voting from team member models,
we arrive at the final prediction.

We evaluate with held-out test sets from the labeled subset of datasets.
Specifically, we conduct three evaluations. 
For each of the 3 labeled datasets VehicleColors, CrawledVehicles, 
and VeRi, we use one for testing and the remaining 2 for building the team.
Results are provided in \autoref{tab:colorcm}.

On each held-out dataset, initial accuracy is $\sim$0.86.
With cross-validation early stopping, we can increase this to $\sim$0.91 
With JPEG compression with 3 ratios, we can increase accuracy by a further 3\%.
By teaming several models, we further increase accuracy to $\sim$0.95
Finally, we add the agreement constraint, where we accept a label only If
$>50\%$ of models have agreed on the label.
This improves labeling accuracy by an additional 3\%, to 0.98. 
On the held-out test set, we can thus label 90\% of the samples, with the 
other 10\% remaining unlabeled due to disagreements.

With these strategies, we label color for BoxCars116K and Cars196 using \corr. 
\sys can label 88\% of the images in these datasets. 
\autoref{fig:corrimages} shows a sample of these images and their assigned labels.

\PP{Type Corroborative Model (Type-CM)}
The team for type annotation labeling is trained with ground truth in 
CompCars, CrawledVehicles, and VeRi to label BoxCars116K and Cars196. 
We use a team of 3 Type-CM models. 
The training process is similar to Color-CM.

\autoref{tab:typecm} shows held-out accuracy on test sets. 
The initial accuracy is 0.87, and with early stopping and JPEG compression, 
we can increase accuracy by 4\%, to 0.94. 
Team of models increases accuracy to 0.96. 
By adding JPEG compression agreement to team members, we arrive at a final accuracy of 0.98.
We can label 93\% of held-out samples with this process.
On our desired ground truth of BoxCars116K and Cars196, the team of Type-CM models 
corroboratively labels 91\% of samples.

\begin{table}[]
    \caption{\textbf{Make-CM}: Accuracy (in mAP) of make classification. Unlike color and type, makes are different across each vehicle. This effectively converts the problem to one of clustering, similar to vehicle re-id. So, we evaluate directly using the feature output and applying conventional re-id evaluation measures.}
    \label{tab:makecm}
    \begin{tabular}{@{}rrrr@{}}
    \toprule
    \multicolumn{1}{l}{}        & CompCars & BoxCars116K & Cars196 \\ \midrule
    \multicolumn{1}{l}{Initial} & 0.73     & 0.69        & 0.65    \\ \midrule
    +Bootstrap                  & 0.76     & 0.71        & 0.68    \\
    +Early Stop                 & 0.77     & 0.73        & 0.70    \\
    +Compression (90)           & 0.79     & 0.74        & 0.72    \\
    +Compression(90, 70, 50)    & 0.79     & 0.74        & 0.73    \\
    +Dynamic Weights            & 0.79     & 0.74        & 0.73    \\
    +Agreement Threshold        & 0.82     & 0.76        & 0.76    \\ \bottomrule
    \end{tabular}
    \end{table}

\PP{Make Corroroborative Model (Make-CM)}
The make annotation labeling team is trained with ground truth in 
CompCars, BoxCars116K, and Cars196. 
We use a team of 3 Make-CM models, with training process similar to Color-CM
and Type-CM.
The evaluation differs, however. 
Since makes are distinct across datasets, we evaluate with 
the mAP metric common in vehicle and person re-id \cite{reidsurvey}.
Specifically, the Make-CM members learn to cluster similar
makes together.
As in re-id, training with batched triplet mining forces backbones
to cluster similar identities together and dissimilar identities further away.
In case of Make-CM, identities are the vehicle makes.
To evaluate, we use mAP metric to check clustering and retrieval accuracy.

\autoref{tab:makecm} shows held-out accuracy on test sets. 
The initial accuracy has more variance, due to difficulty of re-id clustering and
feature learning compared to classification.
On CompCars, we can achieve mAP 0.73, and with early stopping and JPEG compression, 
we can increase accuracy by ~10\%, to 0.82. 
Remaining augmentations only negligible improve mAP across the board.
We can label 80\% of held-out samples by including agreement;
this performance is similar to accuracy in \cite{makecc1, makecc2}.

\subsection{Complementary Features}
\label{sec:compeval}

\begin{figure}[t] 

        \centering
        \includegraphics[width=3.4in]{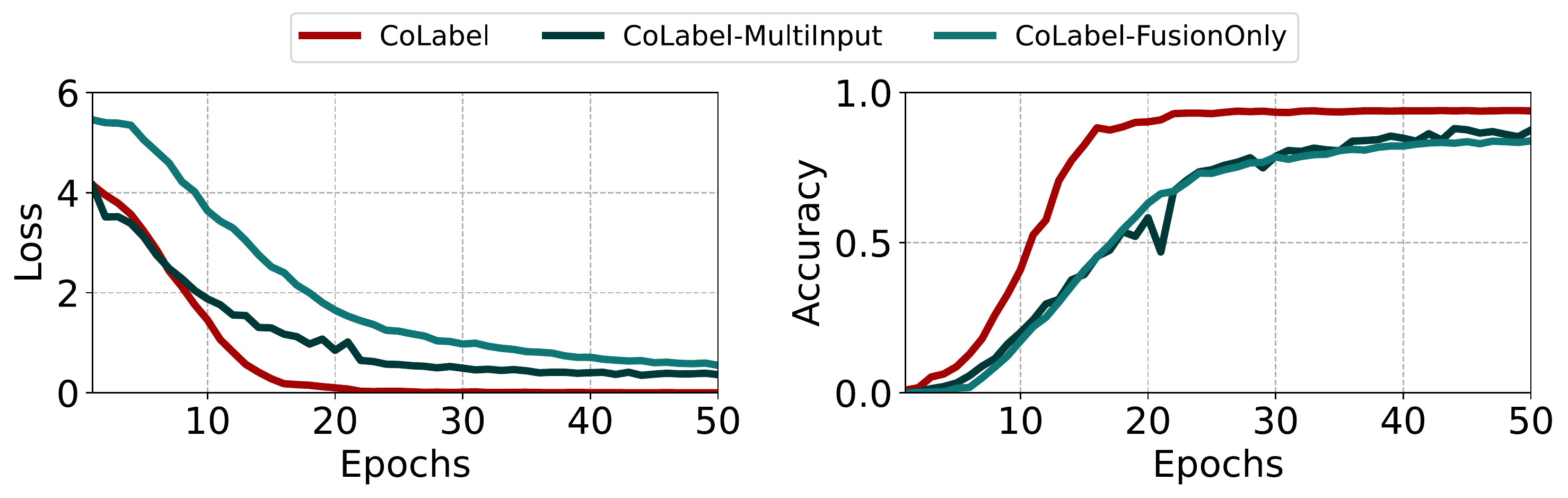}

    \caption{\textbf{Convergence of \sys}: We compare convergence
  with respect to loss minimization and accuracy of \sys against \sysmulti and 
  \sysfusion on CompCars. \sysmulti has multiple inputs, which slows learning of
  fuse-able branch features. \sysfusion uses only the 
  fused feature loss without local harmonization, which reduces effectiveness
  of feature fusion.
   }
  \label{fig:convergence}
\end{figure}

\begin{figure}[t] 

    \centering
    \includegraphics[width=3.4in]{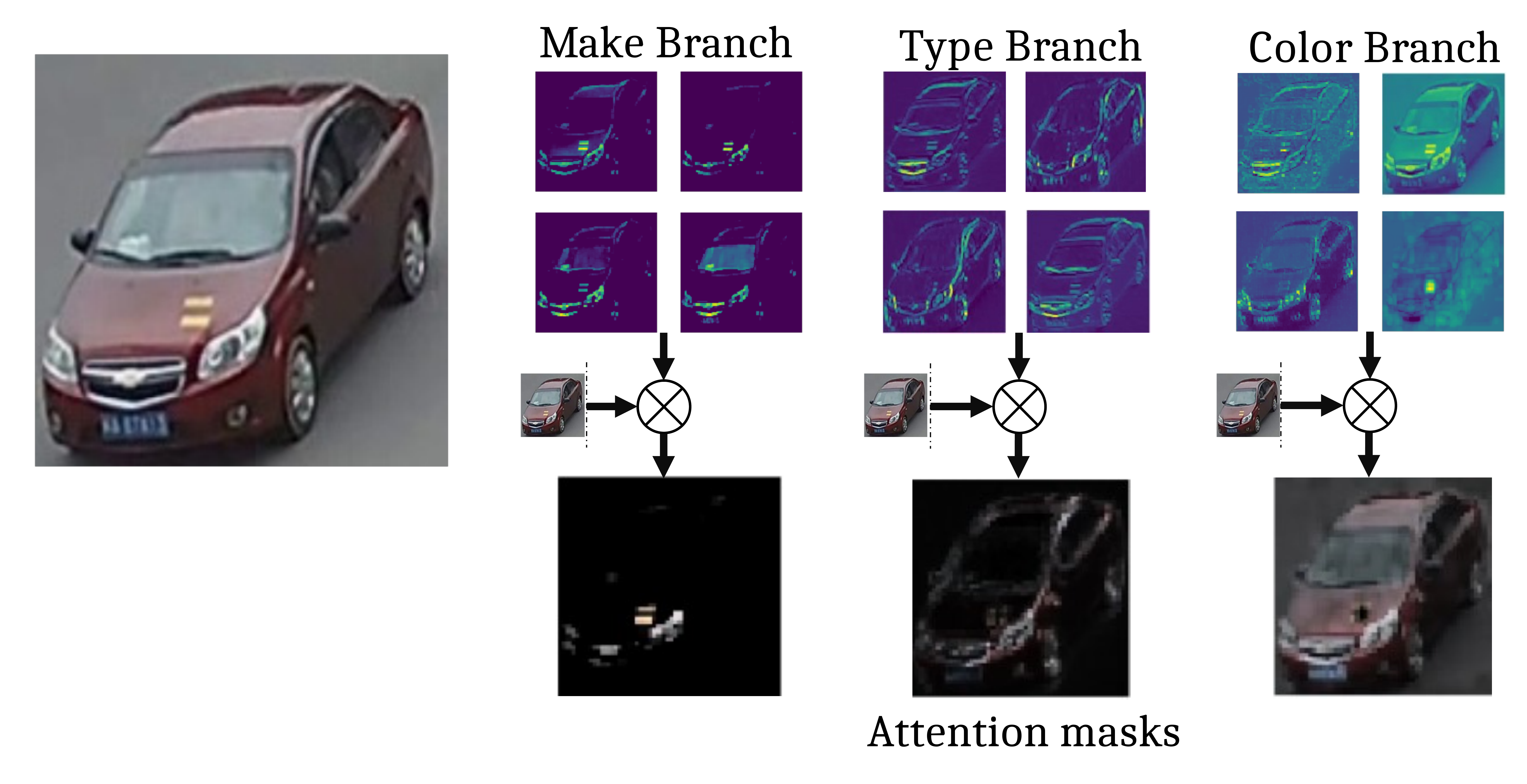}
\caption{\textbf{Attention masks}: We obtain attention masks from the attention modules
in \corr. We show a few of the attention masks in shallow layers of each branch that help
the branch extract their annotation-specific features.
}
\label{fig:attentionmasks}
\end{figure}

After \corr, we have our desired interpretable annotations in the BoxCars116K, 
CompCars, and Cars196 datasets. 
Here, we demonstrate interpretable classification with \comp.

\PP{Shared Input Block}
We first cover the shared input block. 
The shared block is important for feature harmonization to ensure complementary 
features are extracted from a common set of convolutional features to improve
semantic agreement. 
Further, the shared block is updated with backpropagated loss from all branches, 
ensuring the shallow features it extracts are usable by all branches. 
In turn, this improves convergence and training time for \sys. 
We show in \autoref{fig:convergence} the impact of the shared input block
in convergence by comparing loss over time between \sys and \sysmulti, a model
with an input for each branch.
The shared input block improves weight convergence and training time; we
will discuss \sysfusion in the figure in \autoref{sec:colleval}


\begin{table}[t]
    \footnotesize
    \caption{\textbf{Impact of attention}: \sys with attention outperforms model without attention (\sysnoatt) 
    across all classification tasks. For VMMR, attention yields between 4-5\% improvement across all datasets.
    Color accuracies are in parenthesis because these are evaluated on labels generated by Color-CM models, instead of 
    oracle ground truth labels.}
    \label{tab:att}
    \begin{tabular}{l|rr|rr|rr}
    \hline
                   & \multicolumn{2}{c|}{CompCars} & \multicolumn{2}{c|}{Cars196} & \multicolumn{2}{c}{BoxCars116K} \\
                   & CoLabel       & NoAtt      & CoLabel      & NoAtt      & CoLabel        & NoAtt       \\ \hline
    Classification & 0.96          & 0.89          & 0.94         & 0.91          & 0.89           & 0.84           \\ \hline
    Color          & (0.98)             & (0.98)             & (0.98)           & (0.97)             & (0.98)            & (0.98)           \\
    Type           & 0.96          & 0.92          & -            & -             & -              & -           \\
    Make           & 0.95          & 0.91          & 0.92         & 0.87          & 0.87           & 0.81           \\ \hline
    \end{tabular}
\end{table}

\PP{Attention Modules}
\comp also uses attention modules in both the shared input block and the complementary feature branches. 
Attention improves feature extraction in both cases. 
For the shared input layer, attention masks identify image region containing relevant features for each branch. 
For complementary features, attention further improves feature extraction accuracy, shown in ~\autoref{tab:att}.
The make branch improves classification accuracy from 0.91 without attention (\sysnoatt) to 
0.95 by including attention in the branch.

\PP{Attention Masks}
We show attention masks of each branch in \autoref{fig:attentionmasks}. 
While attention masks are generally black-boxes, \sys{'s} interpretability allows us to make
educated guesses about the masks.
Masks for each branch are visually similar to each other, indicating attention has been clustered
by \corr. 
Further, the make branch masks indicate the branch focuses on the logo area of vehicles.
Similarly, the type branch masks focus on the general shape of the vehicle at the edges.
The color branch masks extract overall vehicle color information.

\subsection{Collaborative Learning}
\label{sec:colleval}

\begin{table}[t]
    \caption{\textbf{Impact of Harmonization Loss}: Across all datasets, using both local harmonization loss and final fused loss significantly improves accuracy. }
    \label{tab:conv}
    \begin{tabular}{lrrr}
        \toprule
               & CompCars & Cars196 & BoxCars116K \\ \hline
    CoLabel    & 0.96     & 0.94    & 0.89        \\
    FusionOnly & 0.87     & 0.84    & 0.81      \\ \bottomrule
    \end{tabular}
    \end{table}

Finally, \sys fuses complementary features to generate final features for VMMR classification. 
We evaluate \sys end-to-end to demonstrate the feasibility of inherently interpretable models
with several experiments. 

\PP{Impact of Loss Functions}
First, we show the impact of loss functions on training convergence and accuracy 
on the CompCars dataset.
\autoref{fig:convergence} compares \sys to a \sysfusion, which uses only the final fused loss
and without the local harmonization losses. 
We also show accuracy across datasets in \autoref{tab:conv}.
By adding the local harmonization loss to improve feature fusion, we
can increase accuracy by almost 10\% on average; on CompCars, we increase
accuracy from 0.87 to 0.96 for VMMR.
Without the harmonization loss, \sysfusion converges slower and has lower accuracy.

\begin{table}[t]
    \caption{\textbf{\sys vs Non-Interpretable Models}: We show performance of 
    \sys against several states-of-the-art. 
    \sys achieves similar performance with the added benefit of interpretability. 
    Further, interpretability allows us to exploit disagreements between 
    classifications and existing knowledgebases to further improve accuracy. 
    We show this with \sysmatch, a model that self-diagnoses mistakes 
    using existing knowledge about vehicle makes, models, and types.}
    \label{tab:interpretable}
    \begin{tabular}{lrrr}
        \toprule
                  & CompCars & BoxCars116K & Cars196 \\ \hline
    R50-Att  \cite{baseline}       & 0.90     & 0.75        & 0.89    \\
    R152 \cite{vmmrcmp}         & 0.95     & 0.87        & 0.92    \\
    D161-CMP \cite{vmmrcmp}     & 0.97     & -           & 0.92    \\
    R50-CL \cite{cooccur}       & 0.95     & 0.86        & -       \\ \hline
    CoLabel       & 0.96     & 0.89        & 0.94    \\
    CoLabel-Match & 0.97     & 0.93        & 0.96    \\ \bottomrule
    \end{tabular}
    \end{table}

\PP{Interpretability and Accuracy}
Now, we evaluate \sys against several non-interpretable models: 
(a) R50-Att, a ResNet50 backbone with a single branch with IBN and attention \cite{baseline}, 
(b) R152, a ResNet152 with benchmark results from \cite{vmmrcmp}, 
(c) D161-CMP, a DenseNet with channel pooling from \cite{vmmrcmp}, 
and (d) R50-CL, a ResNet50 with unsupervised co-occurrence learning \cite{cooccur}.

For \sys, we use a ResNet34 backbone. 
The first bottleneck block resides in the shared input block.
The remaining three bottleneck blocks are copied to each branch, as described in \autoref{sec:comp}.
For each model, we use image size 224×224, and train for 50 epochs with lr=1e-4, with a batch size of 64. 

Results are shown in \autoref{tab:interpretable}. 
\sys achieves slightly higher accuracy than both R152 and R50-Att, with accuracy of 0.96 on 
VMMR on CompCars. 
For Cars196 and BoxCars116K, \sys achieves accuracy of 0.94 and 0.89, respectively. 
In each case, \sys achieves similar or slightly better performance than existing non-interpretable approaches. 

However, \sys{'s} results are also interpretable, allowing us to further increase accuracy by \textit{retroactive corrections}. 
Given vehicle models and their ground truth types from existing vehicle databases \cite{nhtsa}, \
we can check where \sys{'s} type detection and vehicle model predictions do not agree. 
This occurs when the image is difficult to process, either due to occlusion, blurriness, or 
other artifacts (an example such disagreement in \corr with 2 cars in the same image is shown in \autoref{fig:corrimages}). 
As such, \sys generates conflicting interpretations, which are themselves useful in analyzing the model. 
Using this variation called \sysmatch, we can further increase accuracy solely due to interpretability, 
to 0.97, 0.96, and 0.93 on CompCars, Cars196, and BoxCars116K, respectively. 

\begin{table}[t]
    \caption{\textbf{Single-Branch Multi-Labeling}: With multi-labeling output from a single branch in \sysml, 
    we can maintain interpretability for a single-branch while sacrificing accuracy.}
    \label{tab:smbl}
    \begin{tabular}{@{}lrrr||r@{}}
    \toprule
                 & CompCars & BoxCars116K & Cars196 & Params \\ \midrule
    HML \cite{ava}         & 0.65     & -           & -       & -          \\
    CoLabel-SMBL & 0.91     & 0.84        & 0.89    & 25M        \\
    CoLabel      & 0.96     & 0.89        & 0.94    & 60M        \\ \bottomrule
    \end{tabular}
\end{table}

\begin{table}[t]
    \caption{\textbf{All-v-All vs 2-Stage Cascade}: We compare \sys under all-v-all and 2-stage cascade. With \syscascade, 
    we use a specialized submodel for each make, simplifying the VMMR problem. 
    We can benefit from interpretability by including retroactive correction to further increase VMMR classification accuracy.}
    \label{tab:ava}
    \begin{tabular}{@{}lrrr@{}}
        \toprule
                      & CompCars      & BoxCars116K   & Cars196       \\ \midrule
    D161-SMP          & {\ul 0.97}    & -             & 0.92          \\
    CoLabel (AVA)          & 0.96          & 0.89          & 0.94          \\
    CoLabel-Match     & 0.97          & 0.93          & {\ul 0.96}    \\
    CoLabel-2SC       & 0.97          & {\ul 0.93}    & 0.95          \\
    CoLabel-2SC-Match & \textbf{0.98} & \textbf{0.94} & \textbf{0.96} \\ \bottomrule
    \end{tabular}
    \end{table}

\PP{Single-Branch Multi-Labeling}
Since \sys uses multiple branches, a natural question is: could branches be removed while 
maintaining interpretability?
We compare \sys{'s} multi-branch interpretability with Single-Branch Multi-Labeling approach (\sysml). 
In \sysml, we use a single branch for feature extraction. 
The features are then used in 4 parallel dense layers: color, type, make, and VMMR detection. 
With \sysml, we could reduce model parameters, since we use a single branch. 
We compare \sysml against \sys and HML \cite{ava} in \autoref{tab:smbl}.
\sysml sacrifices accuracy with the reduced parameters.
Further, we also found \sysml more difficult to converge, as it needed
fine-tuning of learning rates to contend with the multiple backpropagated losses.
We leave further exploration of \sysml with other architecture choices to future work.

\PP{All-v-All vs 2-Stage Cascade}
Here, we evaluate \sys as a 2-stage cascade (\syscascade) and compare to all-v-all (AVA) in \autoref{tab:ava}. 
AVA is the method we have described in \sys, where final features are used for make and model classification. 
In essence, this is a complex problem where \sys{'s} fused features are trained with every vehicle model 
in our datasets. 
In \syscascade, we simplify the problem by creating classifier submodel (\ie a dense layer) for each make. 
The submodels use \sys{'} fused features for prediction. 
So, given the 78 makes in CompCars, we create 78 submodels. 
For each image, \syscascade{'s} vehicle make prediction activates the 
corresponding classification submodel.

Since \syscascade works on a simpler problem, we can improve accuracy 
from \sys, with a trade-off of increase parameters due to the submodels.
\syscascade achieves accuracy of 0.97 on CompCars, 0.95 on Cars196, and 0.93 on BoxCars116K,
comparable to D161-CMP \cite{vmmrcmp}.
With \syscasmatch, we apply \textit{retroactive correction} to further improve 
accuracy on CompCars to 0.98 by verifying predictions
with make-model-type knowledgebase \cite{nhtsa}.

\subsection{Limitations and Future Work}
\label{sec:discuss}
While \sys achieves impressive performance, we have made some assumptions
in its design.
For example, \sys is designed for single-object-per-image.
This is addressed with an object detector such as
YOLO or Mask-RCNN, and 
in case of video streams, a detector UDF over a query engine, similar 
to \cite{chameleon}.
We will conduct a comprehensive ablation study 
to evaluate the impact of interpretable feature branches 
as well as the data domains overlap,
 using recent work on studying high-dimensional dataset overlap
 in \cite{kmp, midas, trust, odin} as a starting point.

\section{Conclusion}
\label{sec:conclusion}

In this paper, we have presented consructive interpretability
with \sys, 
an inherently interpretable model for feature extraction and classification. 
%
%
\dcircle{1} \corr allows us to complete interpretable 
annotations in datasets using a variety of corroborative datasets.
\dcircle{2} \comp perform feature extraction corresponding to 
interpretable annotations, allowing model predictions to be 
inherently interpretable.
Finally, \dcircle{3} \coll lets \sys fuse features effectively
during training using local harmonization losses for each branch.
%
%
%
Our evaluations show that \sys{'s} components make an interpretable
model, with comparable accuracy to state-of-the-art black box models.
We are also able to exploit interpretability to self-diagnose
mistakes in classification, further increasing accuracy with 
\sysmatch and \syscasmatch.

\newpage

\bibliographystyle{ACM-Reference-Format}
\bibliography{main}

\end{document}